\documentclass[journal]{IEEEtran} 

\usepackage{rotating}
\usepackage{cite}
\usepackage{amsmath,amssymb,amsfonts}
\usepackage{algorithmic}
\usepackage{graphicx}
\usepackage{textcomp}
\usepackage{xcolor}
\usepackage{hyperref}
\usepackage{booktabs}
\usepackage{graphicx}
\usepackage{float}
\usepackage[normalem]{ulem}
\useunder{\uline}{\ul}{}

\begin{document}

\author{
    Niels Balemans$^{1, 2}$, Ali Anwar$^{1}$ (Member, IEEE), Jan Steckel$^{2, 3}$, and Siegfried Mercelis$^{1}$
    \thanks{This work was supported by the Research Foundation Flanders (FWO) under Grant Number 1S75624N.}
    \thanks{$^{1}$N. Balemans, A. Anwar and S. Mercelis are with IDLab - Faculty of Applied Engineering, University of Antwerp - imec, Sint-Pietersvliet 7, 2000 Antwerp, Belgium
        {\tt\footnotesize 
            \href{mailto:niels.balemans@uantwerpen.be}{niels.balemans@uantwerpen.be}
        }}
    \thanks{$^{2}$N. Balemans and J. Steckel are with Cosys-Lab - Faculty of Applied Engineering, University of Antwerp, 2000 Antwerp, Belgium}
    \thanks{$^{3}$J. Steckel is with Flanders Make Strategic Research Centre, Lommel, Belgium}
}


\title{LiDAR-BIND-T: Improved and Temporally Consistent Sensor Modality Translation and Fusion for Robotic Applications}

\maketitle
\begin{abstract}
This paper extends LiDAR-BIND, a modular multi-modal fusion framework that binds heterogeneous sensors (radar, sonar) to a LiDAR-defined latent space, with mechanisms that explicitly enforce temporal consistency. We introduce three contributions: (i) temporal embedding similarity that aligns consecutive latent representations, (ii) a motion-aligned transformation loss that matches displacement between predictions and ground truth LiDAR, and (iii) windowed temporal fusion using a specialised temporal module. We further update the model architecture to better preserve spatial structure. Evaluations on radar/sonar-to-LiDAR translation demonstrate improved temporal and spatial coherence, yielding lower absolute trajectory error and better occupancy map accuracy in Cartographer-based SLAM (Simultaneous Localisation and Mapping). We propose different metrics based on the Fréchet Video Motion Distance (FVMD) and a correlation-peak distance metric providing practical temporal quality indicators to evaluate SLAM performance. The proposed temporal LiDAR-BIND, or LiDAR-BIND-T, maintains modular modality fusion while substantially enhancing temporal stability, resulting in improved robustness and performance for downstream SLAM.
\end{abstract}

\begin{IEEEkeywords}
Autonomous Vehicle Navigation, Deep Learning Methods, Robust/Adaptive Control, SLAM, Sensor Fusion
\end{IEEEkeywords}

\section{Introduction}
Reliable perception of the surrounding environment is a fundamental requirement for autonomous systems, such as self-driving vehicles and mobile robots \cite{wang_survey_2025}, \cite{li_multi-sensor_2025}. These systems typically employ a suite of optical sensors, namely cameras and LiDAR (Light Detection and Ranging), to construct a high-fidelity representation of their surroundings. While these sensors provide rich and dense data under ideal circumstances, their performance can be severely compromised by adverse optical environmental conditions like fog, rain, snow, or smoke. Consequently, a strong reliance on only these optical sensing modalities renders these systems vulnerable to critical perception errors and navigational failures, risking safety and operational reliability.

Given the shared vulnerabilities of optical sensors, the integration of additional sensing modalities, such as mmWave radar and sonar, to increase the robustness and accuracy of perception systems is beneficial. Owing to the differences in their physical principles and wavelengths, the signals emitted by these sensors interact with the environment in different ways. By incorporating such diverse sensor types, the system's operating range is broadened and the conditions under which failures may occur are significantly reduced, as these alternative modalities are generally less susceptible to environmental factors that typically affect optical sensors. However, fusing data from such heterogeneous sensors remains challenging due to fundamental differences in characteristics such as spatial resolution, field of view, update rates and intrinsic modality properties. For instance, LiDAR provides precise and dense depth measurements in favourable optical conditions, but its performance deteriorates significantly in adverse optical environments \cite{zhang_perception_2023}. In contrast, radar is more robust under poor optical conditions, yet lacks the capability to deliver dense and high-resolution environmental measurements.

\begin{figure}[t]
    \centering
    \includegraphics[width=\columnwidth]{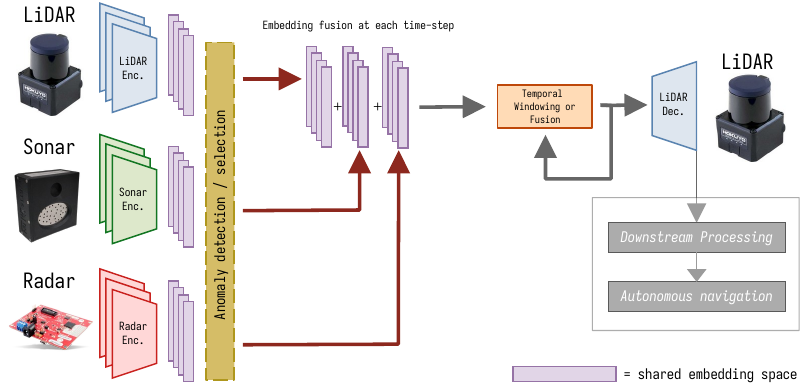}
    \caption[LiDAR-BIND Temporal Framework Overview]{Overview of the improved LiDAR-BIND framework for enhancing temporal consistency in modality translation and fusion. The updated framework fuses the different modalities over multiple time-steps, allowing for more accurate and stable predictions over time.}
    \label{fig:LiDARBIND_temporal_overview_detailed}
\end{figure}

\begin{figure*}[t]
    \includegraphics[width=\textwidth]{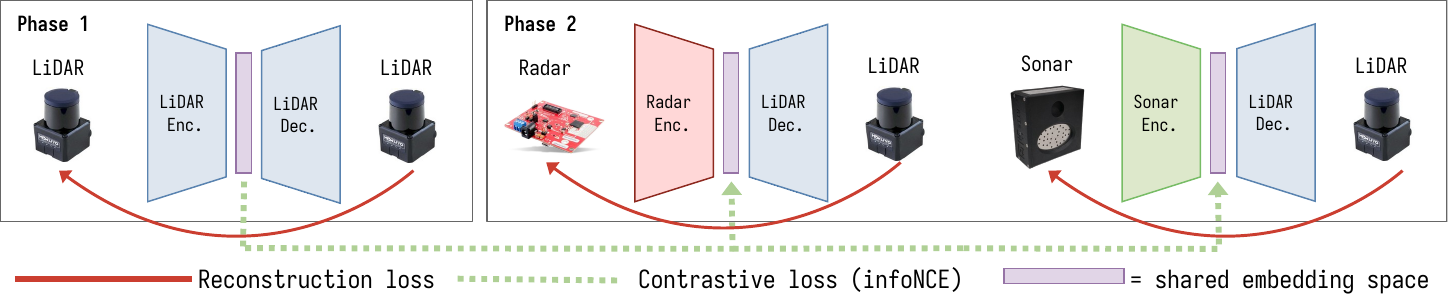}
    \caption{Overview of the LiDAR-BIND framework for multi-modal sensor fusion \cite{balemans_lidar-bind_2024}. The framework aligns embeddings from different sensing modalities into a shared latent space, allowing for accurate and robust data fusion.}
    \label{fig:LiDARBIND_temporal_lidarbind_overview}
\end{figure*}

Previous work proposed several techniques to increase the robustness of existing navigation frameworks by incorporating radar and sonar as additional modalities using novel fusion methods, thereby significantly improving performance under adverse optical environmental conditions \cite{balemans_r2l-slam_2023}, \cite{balemans_s2l-slam_2021}. By leveraging a deep learning-based approach, this method mapped measurements from one sensor modality to another, enabling the translation between modalities. In other words, the goal is to predict how one sensor modality should have perceived the environment based on the measurements acquired by a different sensor. Two distinct and separate modality translation models were proposed: one to predict 2D LiDAR data based on mmWave radar measurements and another to predict LiDAR data from ultrasound readings, employing the eRTIS ultrasound sensor \cite{Kerstens2019}.
During deployment, the system can dynamically choose between using the original LiDAR data, which may be unreliable under adverse optical conditions, or the predictions generated by these models, which are comparatively more robust to degraded optical conditions. This selection-based fusion framework increases the overall robustness and safety of the perception system and enables effective operation in scenarios where LiDAR data alone would be insufficient. However, the selection-based approach limits information fusion by forcing a choice between sensor inputs at each timestep. Making the robustness improvements dependent on the algorithm or method choosing the input sensor at each timestep, prevents full use of complementary sensor strengths and potentially reduces performance in challenging conditions.

To address these limitations, the work in \cite{balemans_lidar-bind_2024} proposed a learning-based framework that establishes a shared latent embedding space across different perception modalities, thereby enabling seamless sensor data fusion within this learned domain. The method employs dedicated encoders for each sensor type to transform modality-specific inputs into feature embeddings; a decoder is then trained to reconstruct 2D LiDAR data from this cross-modal embedding space. This framework, named LiDAR-BIND, highlighting its use of LiDAR data as a binding modality for establishing the shared embedding space, eliminates the need for selection-based approaches. 

While this LiDAR-BIND approach offers notable improvements, it still presents a limitation: namely, it does not explicitly address the consistency of sensor data across time. The absence of temporal continuity is important for the performance in downstream tasks such as Simultaneous Localisation and Mapping (SLAM), where accurate and consistent LiDAR measurements are essential for pose estimation and environment mapping.

The challenge we address in this paper is to produce temporally consistent poses using SLAM and the (fused) LiDAR-BIND predictions. This problem can be linked to a well-recognised challenge in machine learning, particularly in the field of generative modelling, where predictions may often lack coherent evolution over time. Although state-of-the-art methods exist to address temporal inconsistency, applying them in this multimodal fusion context is not straightforward, and has not been attempted before. These techniques are typically optimised or trained for real-world RGB images and, as a result, perform suboptimally when used with non-RGB sensor modalities.

In this paper, we propose improvements to the LiDAR-BIND framework that enable the fusion of multiple time steps across different sensor modalities, thereby improving both temporal and spatial consistency in predictions. These improvements to the predictions contribute to the main goal of better performance in downstream tasks such as Simultaneous Localisation and Mapping (SLAM). The main contributions of this work are as follows:
\begin{itemize}
    \item We introduce methods for improving the temporal quality of sensor data predictions in the LiDAR-BIND framework. This method leverages a temporal windowing and fusion mechanism and specialised losses to enforce consistency across consecutive frames, leading to improved LiDAR reconstruction performance.
    \item We adapt and propose relevant metrics to evaluate the temporal consistency of predicted sensor data. This includes a metric based on the transformation between measurements and an adapted version of the Frechet-Video-Motion distance metric (FVMD), proposed in \cite{liu_frechet_2024}. We show that a lower FVMD score corresponds to better temporal consistency, which in turn leads to improved SLAM performance.
    \item We evaluate the performance of the proposed methods on SLAM tasks, demonstrating that the enhanced temporal consistency leads to more accurate and reliable navigation, particularly in challenging conditions where optical data is degraded or unavailable.
\end{itemize}

This paper is organised as follows. We start with a review of state-of-the-art approaches for temporal consistency in deep learning and their relevance to multi-modal sensor fusion in Section~\ref{sec:LiDARBIND_temporal_consistency_sota}. Next, Section~\ref{sec:LiDARBIND_temporal_metrics_sota} discusses the typical metrics for temporal consistency and their limitations for the non-RGB sensor measurement predictions. Section~\ref{sec:LiDARBIND_temporal_lidarbind} describes our hypothesis on the temporal consistency problem observed in LiDAR-BIND and the challenges of improving the temporal quality of predictions using the shared latent space. In Section~\ref{sec:LiDARBIND_temporal_method}, we present the improvements to the LiDAR-BIND framework. Section~\ref{sec:LiDARBIND_temporal_experiments} details the experimental setup and results. Section~\ref{sec:LiDARBIND_temporal_ablation} provides an ablation study of the proposed components. Finally, Section~\ref{sec:LiDARBIND_temporal_discussion} discusses the results and outlines directions for future work.

\section{Consistency in Predictions: Inspiration from State-of-the-Art}
\label{sec:LiDARBIND_temporal_consistency_sota}
Achieving temporal consistency in deep learning models is a central challenge across fields such as robotics, control systems and image and video generation. As many state-of-the-art methods have been developed in these fields, their core principles provide valuable inspiration for improving prediction consistency in the LiDAR-BIND multi-modal sensor fusion framework. However, direct application of these methods to the sparse and heterogeneous sensor data often encountered in robotics (e.g., radar, sonar, LiDAR), as shown in Figure \ref{fig:LiDARBIND_temporal_robot}-a, is typically non-trivial.
Here, we highlight key state-of-the-art approaches and how their ideas inform our proposed improvements. We present these techniques in the categories we observed most prominently across these domains: (1) learning temporally consistent representations, and (2) efficient (long-term) temporal modelling.

\textbf{Learning Temporally Consistent Representations:}
A recurring theme in state-of-the-art is the explicit enforcement of temporal relationships in the latent space of the model to ensure consistent predictions across time. For example, Video Depth Anything (VDA)~\cite{chen_video_2025} introduces a temporal gradient matching loss that enforces consistency in depth changes between adjacent frames, without relying on optical flow or geometric priors. Similarly, the temporally-consistent autoencoders for control systems, as presented in \cite{chakrabarti_temporally-consistent_2025,somma_hybrid_2025}, use a regularisation in the latent space, aligning time derivatives of latent variables with their finite difference approximations. Temporal Masked Autoencoders (T-MAE)~\cite{wei_t-mae_2024}, designed for LiDAR point clouds, reconstruct masked portions of a scan using information from previous frames, enabling the model to learn temporal dependencies and robust representations.

The recent advances in image and video generation further highlight the importance of temporally and spatially consistent representations. For example, MoVideo, presented in \cite{liang_movideo_2024}, improved the temporal quality of generated videos by proposing a motion-aware framework that explicitly takes into account video depth and optical flow. This approach, along with methods such as ConsistI2V \cite{ren_consisti2v_2024} and LATTE (Latent Diffusion Transformer) \cite{ma_latte_2025}, underscores the significance of both spatial-temporal modelling within the latent representation space, a technique we will incorporate in our proposed improvements. Additionally, the method presented in \cite{zhang_training-free_2025} introduces a motion consistency loss, further demonstrating the benefits of accounting for underlying motion in video generation.

In this work, we extend and adapt the LiDAR-BIND framework to achieve better spatio-temporal consistency within the embedding space. This is accomplished by encouraging similarity between embeddings of consecutive measurements and implementing architectural modifications, which are further detailed in Section~\ref{sec:LiDARBIND_temporal_method}. Our approach is inspired by the objective of learning robust representations capable of capturing temporal dependencies. By aligning the embeddings of consecutive frames, we ensure that the model learns to produce embeddings that are coherent, but not identical, for temporally adjacent frames, thereby achieving both temporal smoothness and sensitivity to actual changes in the environment.

\textbf{Efficient and Scalable Temporal Modelling: }
In addition to ensuring the consistency of embeddings between timesteps, modelling long-term temporal dependencies is crucial for our applications. TECO \cite{yan_temporally_2023} introduces a transformer-based architecture that compresses video data into discrete embeddings and models long-term dependencies using a temporal transformer, thereby enabling scalable and consistent video prediction. VideoLCM~\cite{wang_videolcm_2023} leverages consistency models to synthesise high-quality videos with minimal sampling steps, demonstrating that efficient temporal modelling can be achieved without sacrificing quality. These approaches are conceptually aligned with our proposed use of temporal windowing and fusion modules, as they allow the model to aggregate information across multiple timesteps and effectively filter out anomalies.

In this work, we adapt these principles to the multi-modal sensor fusion context, tailoring our improvements to address the unique challenges inherent in combining data from heterogeneous sensors. While a direct adoption of existing video-based methods is not feasible due to differences in data modality and task objectives, our approach draws on the underlying strengths of these techniques, such as robust temporal modelling and aggregation of information, to enhance the temporal consistency and robustness of LiDAR-BIND predictions.

\section{Metrics for Temporal Consistency and Their Limitations}
\label{sec:LiDARBIND_temporal_metrics_sota}

While investigating state-of-the-art methods to improve temporal consistency, we also examined the metrics typically used to evaluate these improvements. Several metrics have been developed for video generation and processing, but they often present limitations when applied to the sparse and heterogeneous sensor data of our application.

\begin{itemize}
    \item \textbf{Fréchet Video Distance (FVD) \cite{unterthiner_towards_2019}}: This is one of the most widely used metrics, comparing the distribution of video features extracted using a pre-trained network between the generated and reference video sets. While effective at capturing general temporal coherence, FVD tends to be more sensitive to image fidelity rather than long-term consistency in motion patterns.
    \item \textbf{FID-VID (Fréchet Inception Distance for Videos) \cite{heusel_gans_2018}}: This extends the image-based FID by assessing generated frames using a pre-trained image classification model, primarily focusing on the visual quality of individual frames rather than their temporal relationships.
    \item \textbf{Frame-level Metrics (PSNR, SSIM)}: Peak Signal-to-Noise Ratio (PSNR) and Structural Similarity Index Measure (SSIM) are often applied to video data, but they are primarily applied to single timesteps and are therefore less effective at evaluating temporal consistency across frames.
    \item \textbf{VBench \cite{huang_vbench_2023, huang_vbench_2024}}: This provides a comprehensive suite to evaluate video generation models, the temporal quality metrics include subject consistency, background consistency, temporal flickering and motion smoothness, and relies heavily on pre-trained models to extract the relevant metric features.
    \item \textbf{Fréchet Video Motion Distance (FVMD) \cite{liu_frechet_2024}}: To address the limitations of existing metrics, particularly concerning complex motion patterns, FVMD was proposed as a novel metric focusing explicitly on motion consistency in video generation. FVMD measures temporal motion consistency based on the patterns of velocity and acceleration in video movements, extracting motion trajectories of key points using a pre-trained point tracking model like PIPs++ \cite{zheng_pointodyssey_2023}. This metric effectively detects temporal noise and aligns better with human perceptions of generated video quality than existing metrics, even in the presence of intensive motion.
\end{itemize}

In our application, conventional evaluation metrics face significant limitations due to the sparse and heterogeneous characteristics of our data, as illustrated in Figure \ref{fig:LiDARBIND_temporal_robot}-a. Many of these standard tests rely on pre-trained image classification or object detection models for typical camera imagery. However, these approaches are not applicable to our context, where the data does not consist of standard images but rather originates from our other sensor modalities. Therefore, there is a clear need for domain-specific evaluation methods that can effectively assess the relevant aspects of consistency within our dataset. In this work, we present an attempt at creating metrics evaluating the temporal quality of our predictions for our specific use case of multi-modal sensor fusion for SLAM.

\section{LiDAR-BIND consistency problem}
\label{sec:LiDARBIND_temporal_lidarbind}
As mentioned previously, LiDAR-BIND \cite{balemans_lidar-bind_2024} proposed a framework for multi-modal sensor fusion that employs a shared embedding space across different perception modalities. Next to its ability to seamlessly fuse multiple modalities comes the benefit that there are no requirements for a large synchronised dataset across modalities; instead, the sole requirement is samples of each desired modality combined with the binding modality, which in our case is LiDAR. The framework operates through a two-phase training process, presented in Figure \ref{fig:LiDARBIND_temporal_lidarbind_overview}. In the first phase, a transformer-based autoencoder is trained to define a latent vocabulary with the binding modality (LiDAR) serving as the reference. Subsequently, in the second phase, the modality-specific encoder/decoder networks are trained for sensors such as radar and sonar. Their respective embeddings are aligned with the pre-established latent space using a contrastive loss function (infoNCE \cite{Oord2018}) combined with cosine similarity as the metric.

\begin{figure}[t]
    \centering
    \includegraphics[width=\columnwidth]{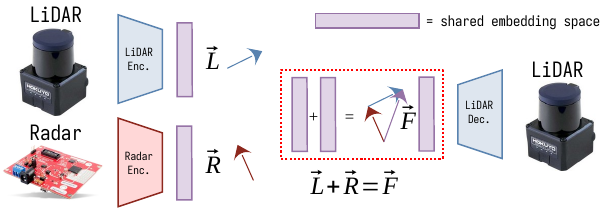}
    \caption{LiDAR-BIND latent fusion mechanism. The framework aligns embeddings from different sensing modalities into a shared latent space, allowing for accurate and robust data fusion. Note that this figure exaggerates the angular orientation of the vectors for illustrative purposes. In normal conditions, these vectors will be encoded closer to each other, while anomalies can be detected by a significant deviation in the angular orientation.}
    \label{fig:LiDARBIND_temporal_latent_fusion}
\end{figure}

The major innovations of LiDAR-BIND lie in its approach to latent space fusion. The utilisation of cosine similarity forces the important information in the angular orientation of each embedding vector in the shared latent space. This property allows for arithmetic combination, such as simple vector addition, depicted in Figure~\ref{fig:LiDARBIND_temporal_latent_fusion}, of embeddings from different modalities, resulting in a fused representation that captures environmental context from all sensor inputs.

Beyond facilitating this multimodal fusion into a robust and unified embedding and measurement prediction, the mechanism enables on-the-fly anomaly detection by comparing the orientations of embeddings produced by each modality. If a particular sensor produces an embedding that notably deviates from those of other modalities, the fusion process can be dynamically adapted to rely more heavily on the reliable modalities, thereby further enhancing system robustness. Additional details regarding the LiDAR-BIND framework and the fusion mechanism can be found in earlier work~\cite{balemans_multi-modal_2024}.

While LiDAR-BIND demonstrated significant improvements in robustness under degraded optical conditions, the lack of temporal consistency remains a limitation that constrains the performance and accuracy of downstream applications such as SLAM, specifically when heavily relying on scan matching.
Compared to the individual CNN-based modality translation models, which served as a baseline for the framework, LiDAR-BIND appears more susceptible to temporal inconsistency due to its reliance on (vision) transformers \cite{Dosovitskiy2020} and full-rank matrix multiplication as learnable linear operators. These components, although advantageous in delivering encoding performance and enabling seamless multimodal fusion in the embedding space (as demonstrated previously), do not directly preserve spatial relationships. This lack of support to maintain spatial features through the encoding process makes it especially challenging to produce consistent predictions when processing only a single frame at a time.

We believe this issue can be addressed by incorporating methods that fuse information from multiple consecutive timesteps, while still keeping computational requirements low and retaining the benefits of the framework, namely efficient low-level fusion of sensor measurements. To this end, we draw inspiration from state-of-the-art techniques in deep learning designed to boost temporal embedding alignment. Additionally, we modify the architecture of the model to rely more on convolutions as linear learnable operators instead of the full rank matrix multiplications used in the previous work, which helps preserve spatial information and ensures that local features remain consistent throughout the encoding process.

\textbf{Latent Space Design Considerations: }
Next to this architectural issue, we also hypothesise that the modelling of the latent space itself contributes to the temporal inconsistency observed in LiDAR-BIND predictions. Currently this space is entirely defined by the binding modality (LiDAR), which does not suffer from the same temporal inconsistencies as the other modalities. For example multi-path reflections in both radar and sonar can cause significant variations in measurements between consecutive frames, even when the environment remains largely unchanged. This noise can lead to embeddings that are not only inconsistent over time but also misaligned with the corresponding LiDAR embeddings, which are more stable.
We refer to this latent space as "shared" because it provides a common ground for aligning embeddings from different modalities and time steps. However, this configuration means that the space is only optimised to capture features relevant to 2D LiDAR data. Although this should be sufficient for the decoder to function correctly, it can present challenges for the encoders of other modalities. These encoders may have difficulty extracting relevant features in relation to the desired 2D LiDAR prediction, particularly since the other modalities are sparser and more affected by sensor and environmental noise. Such noise inevitably degrades the quality of the fused representation.

In contrast, the baseline CNN models did not use a phased training approach and used an embedding space optimised for each separate modality. To investigate the effects of different latent space designs, we present an experiment to analyse the impact of different latent configurations on model performance as well as its ability to generate consistent predictions. Achieving better prediction consistency is expected to lead to improved SLAM performance.

\begin{figure*}[ht]
    \centering
    \includegraphics[width=\textwidth]{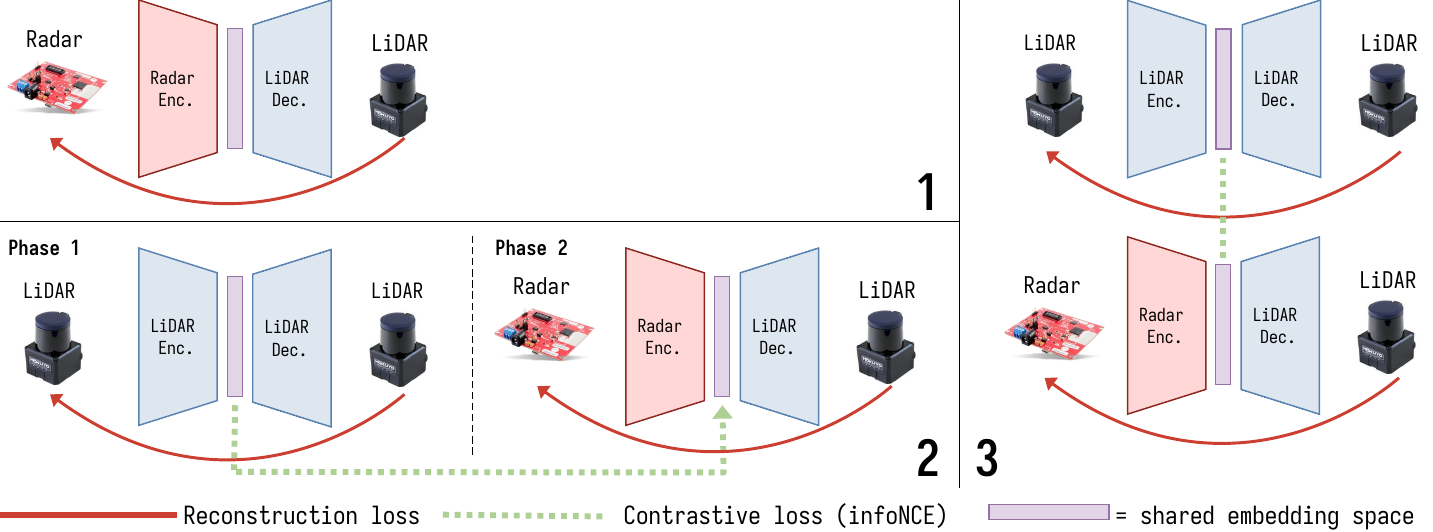}
    \caption[LiDAR-BIND Temporal Latent Space Experiment Overview]{Overview of the different embedding space configurations tested in our experiments. (1) The first configuration is defined using radar data, (2) in the second configuration, the latent space is defined by the LiDAR data (similar to LiDAR-BIND), and (3) the third configuration combines both LiDAR and radar modalities to define the latent space at the same time. With this test, we aim to investigate the impact of embedding space design on model performance and prediction consistency.}
    \label{fig:LiDARBIND_temporal_latent_space_test}
\end{figure*}

For this test, we created three versions of a R2L (radar to LiDAR) model, ensuring that each used the same encoder and decoder architectures. The variations involved only the method of defining the embedding space. In one version, the latent space was defined using radar data. In another version, it was defined by LiDAR data, similar to the LiDAR-BIND framework. For the third version, both LiDAR and radar modalities were combined to define the latent representation. Figure~\ref{fig:LiDARBIND_temporal_latent_space_test} provides an overview of these versions and the corresponding training procedures. The same embedding alignment losses used in LiDAR-BIND were applied, together with modality-specific reconstruction losses.
SLAM results were obtained using Google Cartographer \cite{noauthor_cartographer_nodate}, which was configured to heavily rely on scan matching instead of wheel encoder odometry data in determining the transformation between scans. With this setup, models that produce more temporally consistent predictions perform better and result in lower absolute positional error (APE). The experimental results are presented in Table~\ref{tab:LiDARBIND_temporal_latent_space_ablation}.

\begin{table}[ht]
    \centering
    \caption[Embedding space ablation study]{Absolute positional errors (APE) in meters for the different latent space configurations along our validation sequences in good optical conditions.}
    \label{tab:LiDARBIND_temporal_latent_space_ablation}
    \begin{tabular}{l|ccc}
        \toprule
        & \multicolumn{3}{c}{\textbf{APE SLAM results}} \\
        \textbf{Latent Configuration} & RMSE & MEAN & STD \\
            \midrule
 Radar latent      & 0.087 & 0.075 & 0.043 \\
 LiDAR latent      & 1.710 & 1.432 & 0.935 \\
 Combined latent   & 0.133 & 0.116 & 0.065 \\
        \bottomrule
    \end{tabular}
\end{table}

Examining these results, we observe that the radar-only latent space configuration outperforms the other configurations in terms of absolute positional error (APE). This suggests that, because the embedding space is specifically optimised for radar data, it generates more consistent predictions, which leads to superior SLAM performance, particularly in scenarios where scan matching is critical. We can interpret this as the model learning that different radar-scans, with and without noise (multi-path reflections), should still produce similar embeddings and predictions, as they represent the same scene. This results in more consistent predictions over time, which is crucial for effective SLAM performance. Additionally, this experiment highlights a trade-off inherent to the LiDAR-BIND framework between its modularity and flexibility and the temporal consistency and SLAM accuracy of its predictions. With LiDAR-BIND, it is possible to add new modalities after deployment, simply by training an appropriate encoder, owing to its phased training approach. However, this flexibility may come at the expense of reduced temporal consistency and SLAM performance.

The latent space defined using only LiDAR data allows the seamless fusion of radar and LiDAR in latent space; however, it does not perform as effectively as the radar-only latent space in our experiments. A likely explanation is that, where consecutive radar measurements often vary due to sensor noise, the corresponding LiDAR measurements remain relatively stable. Consequently, the model does not learn that such noise in the measurements should still produce similar embeddings and predictions. This leads to inconsistent predictions over time, thereby degrading SLAM performance.
The configuration in which both LiDAR and radar data are used to define the latent representation demonstrates that optimising the embedding space jointly for both modalities can improve performance. However, this improvement is accompanied by a loss of modularity and flexibility. Specifically, such a configuration would require all desired modalities to be present and synchronised in the training dataset. This constraint makes the emergent cross-modal alignment achieved by the LiDAR-BIND framework—for example, the alignment between sonar and radar—unfeasible in the combined latent space approach.

Given this experiment, we continue with the modular and flexible approach offered by the LiDAR-BIND framework, and aim to improve this framework with the methods proposed in this work, for instance, by incorporating multiple time steps into the fusion process to mitigate measurement noise.

To summarize, we identified that the temporal consistency of LiDAR-BIND generated predictions can be significantly improved. The original framework, while effective at modality translation for individual timesteps, does not explicitly take into account the relationship between consecutive measurements. In addition, architectural and embedding space limitations were identified that lead to temporal discontinuities, particularly when translating from noisy sensor modalities like radar and sonar, where transient effects such as multi-path reflections can cause significant variations between frames. In the next section, we present our proposed improvements to the LiDAR-BIND framework that aim to address these challenges.

\section{Temporal Improvements to LiDAR-BIND}
\label{sec:LiDARBIND_temporal_method}
In this section, we present modifications to the LiDAR-BIND framework that aim to improve the temporal quality of the predicted sensor measurements. The performance and usability of the translated sensor data in common robotic tasks such as SLAM can be significantly affected by the inconsistency between time steps, especially in scenarios where only spatially sparse modalities, such as radar or sonar, are available and no real LiDAR data can be used. These SLAM algorithms depend on the consistency of features across sequential data to accurately track landmarks and estimate movement. If spatial features in the predicted data change or exhibit unpredictable jitter between consecutive frames, the algorithm becomes unable to reliably track landmarks, which results in failure to construct an accurate map and estimate the vehicle's position.

In Section \ref{sec:LiDARBIND_temporal_lidarbind}, we identified that the temporal consistency of LiDAR-BIND generated predictions can be significantly improved. The original framework, while effective at modality translation for individual timesteps, does not explicitly take into account the relationship between consecutive measurements. This leads to temporal discontinuities, particularly when translating from noisy sensor modalities like radar, where transient effects such as multi-path reflections can cause significant variations between frames.

To address this challenge, we introduce three key additions to the framework: (1) enforcing embedding similarity between consecutive measurements, (2) introducing a transformation-based loss for prediction alignment, and (3) implementing temporal windowing and a temporal fusion module. The goal is to better maintain spatial features between consecutive measurements by including more temporal context in the fusion process, and making sure the motion in the input frames matches the motion of the predicted frames. Finally, we updated the architecture of the encoder and decoder to better preserve spatial features within the data. Specifically, we replaced several fully connected layers from the model with convolution layers, as these operations are more effective in capturing local patterns and maintaining spatial hierarchies. Finally, instead of dividing the input measurements into individual patches for processing with the vision transformer, we adopted a convolutional approach to create the transformer embeddings. This modification also contributed to improved preservation of spatial consistency throughout the modality translation and fusion processes.


\subsection{Embedding Similarity Between Consecutive Measurements}
\begin{figure}[t]
    \centering
    \includegraphics[width=\columnwidth]{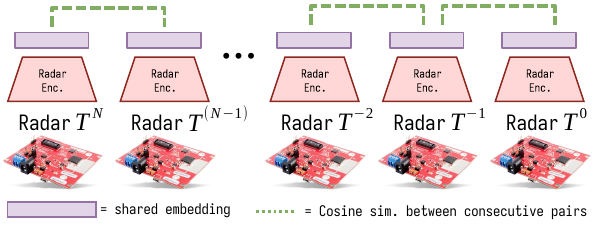}
    \caption[Temporal Embedding Similarity Loss Visualisation]{Visualisation of the embedding similarity loss. Consecutive embeddings are specifically trained to be close in embedding space, promoting consistency in predictions.}
    \label{fig:LiDARBIND_temporal_similarity}
\end{figure}

In the original LiDAR-BIND framework, encoders are trained to map different modalities into a shared latent space, ensuring, for example, that a sonar measurement of a scene would have a similar embedding to the synchronised LiDAR measurement of the same scene. Building on this concept, we extend the training objective to enforce high similarity between embeddings of consecutive measurements across time. This approach assumes that, under normal operating conditions, sensor measurements between time steps are similar, provided the time interval is not too large and that, for a vehicle moving at a certain velocity, the environment as perceived by the sensor will not change drastically between close time intervals. Figure~\ref{fig:LiDARBIND_temporal_similarity} visualises this concept. As input to our models, we do not change from the original LiDAR-BIND framework and use processed range-azimuth heatmaps of the sensor data, an example of which can be seen in Figure \ref{fig:LiDARBIND_temporal_robot}-a. These heatmaps are generated from the raw sensor data and represent the range information in a 2D grid format, where each cell corresponds to a specific azimuth angle and range value.

Similar to the original framework, we employ the cosine similarity to enforce consistency between time steps. For any two consecutive embeddings, $e_t$ and $e_{t-1}$, the similarity is maximized according to the following equation:
\begin{equation} 
    \label{eq:temporal_cos_sim} 
    \mathcal{L}_{\text{sim}}= 1 - \frac{e_t \cdot e_{t-1}}{\|e_t\| \|e_{t-1}\|}
\end{equation}

We employ this loss during the first phase of LiDAR-BIND training, in which the LiDAR encoder and decoder are trained to ensure similarity among consecutive embeddings. Compared to the original LiDAR-BIND, we add here an explicit alignment between the embeddings during the first phase, for which there was previously no constraint on the latent space. We also employ this training objective in the second phase of training, where the modality-specific encoders are trained. However, using this loss in this phase reveals a possible conflict with the contrastive loss (infoNCE) applied at this stage to align cross-modal embeddings. Employing a contrastive loss function is beneficial because it encourages the model to learn more discriminative features between samples. The contrastive objective's effectiveness is dependent on the batch size, as negative samples are defined by the other samples present within the batch. This dependency introduces a conflict, since the time steps we attempt to align for temporal consecutive similarity could also appear as negative samples for the contrastive loss, resulting in potentially contradictory training objectives.

As part of our ablation experiments, we examine the effect of temporal consecutive similarity loss with respect to the contrastive loss. We also experiment with a regularisation within the contrastive InfoNCE loss function, where embeddings of spatially similar inputs (as determined by the LiDAR scans) are penalised less strongly as negative samples. In this way, temporally closely related measurements are encouraged to be nearer in the embedding space, instead of being pushed apart. We still include these measurements as negative samples in order to promote learning of robust and discriminative features, as we do not want close time-steps to generate the same embedding and thus the same LiDAR prediction. The results of this regularisation are also presented in the ablation study, and the other results presented in this paper are with the proposed temporal similarity loss enabled in the second training phase.

\subsection{Prediction Transformation Consistency}
A critical component of SLAM is scan matching, where consecutive measurements are used to estimate the robot's motion. To train for this task specifically, we looked for an optimisation function that would directly promote consistency in the geometric transformation between consecutive predicted frames. While initial experiments explored using odometry data as an additional output for our models, this proved problematic. The dataset often contained long periods of constant velocity, introducing a strong bias. Moreover, changes in odometry do not always correspond to perceptible changes in our processed range-azimuth heatmaps that serve as the only input for our encoders. This mismatch makes it difficult to train the model effectively, as the model can not learn to predict the correct motion based on these heatmaps alone. Using odometry data as an input to inform the model about the expected motion is also not feasible, for the same reason; changes in the odometry do not always correspond to changes in the heatmap, and the fact that our dataset consists largely of the same velocities.

\begin{figure}[t]
    \centering
    \includegraphics[width=\columnwidth]{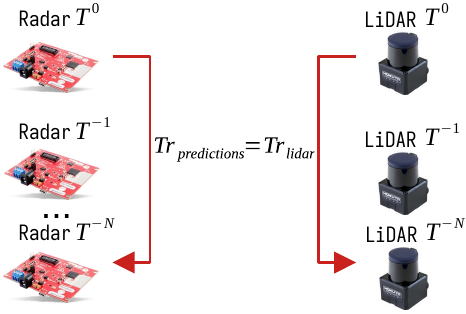}
    \caption[Transformation Consistency Loss Visualization]{Visualization of the transformation consistency loss. The model is trained to align the distribution of likely displacements between predicted frames ($p_{t-1}, p_t$) with the displacement distribution from the ground-truth frames ($l_{t-1}, l_t$).}
    \label{fig:LiDARBIND_temporal_transformer_temporal_loss}
\end{figure}

We, therefore, opted for an approach that relies only on the information visible to the model: the heatmaps themselves. We use 2D cross-correlation to measure the translational shift between two consecutive frames, $t-1$ and $t$. The peak in the resulting correlation map indicates the most likely displacement vector that aligns the two frames. Our objective is to ensure that the displacement distribution between two consecutive predicted frames, $p_t$ and $p_{t-1}$, matches the displacement distribution between the corresponding ground-truth LiDAR frames, $l_t$ and $l_{t-1}$. This concept is visualised in Figure~\ref{fig:LiDARBIND_temporal_transformer_temporal_loss}.

Let $C(A, B)$ be the 2D cross-correlation map of images A and B. To transform these correlation maps into probability distributions over possible displacements, we apply a separable 2D softmax operation. This is performed by applying the softmax function independently along the height and width dimensions and multiplying the results, which strongly sharpens the primary peak. Let this operation be denoted by $\hat{S}(\cdot)$. We then define the transformation consistency loss $\mathcal{L}_{\text{T}}$ using the Kullback-Leibler (KL) Divergence, which measures the difference between the two resulting probability distributions:
\begin{equation}
     \label{eq:correlation_loss} 
     \begin{gathered} 
 Q_p = \hat{S}(C(p_t, p_{t-1})) \quad , \quad Q_l = \hat{S}(C(l_t, l_{t-1})) \\
        \mathcal{L}_{\text{T}} = D_{\text{KL}}(Q_l \,||\, Q_p) 
     \end{gathered}
\end{equation}
Minimising this loss forces the probability distribution of likely displacements for the predicted frames ($Q_p$) to match the ground-truth distribution ($Q_l$). This provides a robust, differentiable signal, enabling the model to learn to generate temporally consistent frame sequences without direct supervision from odometry.

Furthermore, this 2D correlation mechanism serves as a robust evaluation metric, as detailed in Section~\ref{sec:LiDARBIND_temporal_experiments}. By finding the coordinates of the peak ($\operatorname*{arg\,max}$) in the predicted correlation map $Q_p$ and comparing it to the peak location in the ground-truth map $Q_l$, we can quantitatively assess the accuracy of the predicted motion in pixels.

\subsection{Temporal Windowing and Temporal Fusion}
Initial experiments revealed that enforcing embedding similarity and transformation consistency alone was insufficient, particularly for modalities such as radar, where measurements can vary significantly between time steps due to environmental noise and multi-path effects. The original LiDAR-BIND model, which processes one measurement at a time, struggles to filter out such anomalies, as it lacks temporal context. If the source measurements themselves are not temporally consistent, a model operating on a single timestep cannot be expected to produce a consistent output, specifically if that model uses a latent representation space determined by another modality that does not suffer from such measurement inconsistencies.

To address this, we introduce temporal windowing: a sliding window of size $N$ is applied over the sequence of embeddings, analogous to the latent fusion approach used for multi-modal data presented in Section \ref{sec:LiDARBIND_temporal_lidarbind}. Instead of fusing different modalities at a single time step, we now fuse multiple time steps of the same modality in latent space. 

\begin{figure}[t]
    \centering
    \includegraphics[width=\columnwidth]{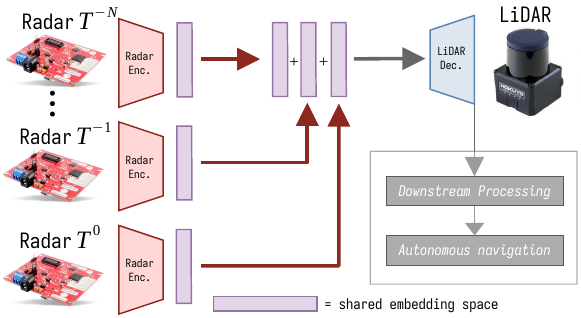}
    \caption[Temporal Windowing Overview]{Overview of temporal windowing. A sliding window over the embeddings is fused using simple addition, essentially averaging the embeddings over this window, before using the LiDAR decoder to produce the final predicted sensor data.}
    \label{fig:LiDARBIND_temporal_overview_temporal_windowing}
\end{figure}

However, simply averaging over a window, as depicted in Figure \ref{fig:LiDARBIND_temporal_overview_temporal_windowing}, can introduce a delay in the predictions, as we can also expect this averaging in the predictions, which worsens as the window size increases. To mitigate this, we incorporate a temporal fusion module that processes the windowed embeddings and outputs a single, temporally fused embedding that is both consistent and responsive to recent changes. We opt for a convolution-based fusion approach where the kernel operates over the time dimension of the stacked latent embeddings. This is to ensure only local spatial features influence the fusion process, allowing the model to focus on relevant temporal information while maintaining spatial coherence. This convolution-based approach, combined with the above-mentioned changes, enables the model to learn complex temporal dependencies and potentially filter out measurement anomalies that would otherwise be undetected by our encoder that operates on a single time step. An overview of this process is shown in Figure~\ref{fig:LiDARBIND_temporal_training_overview}, where this windowing and temporal fusion idea are presented in phase three.

\subsection{Integration into the LiDAR-BIND Framework}
\begin{figure*}[ht]
    \centering
    \includegraphics[width=\textwidth]{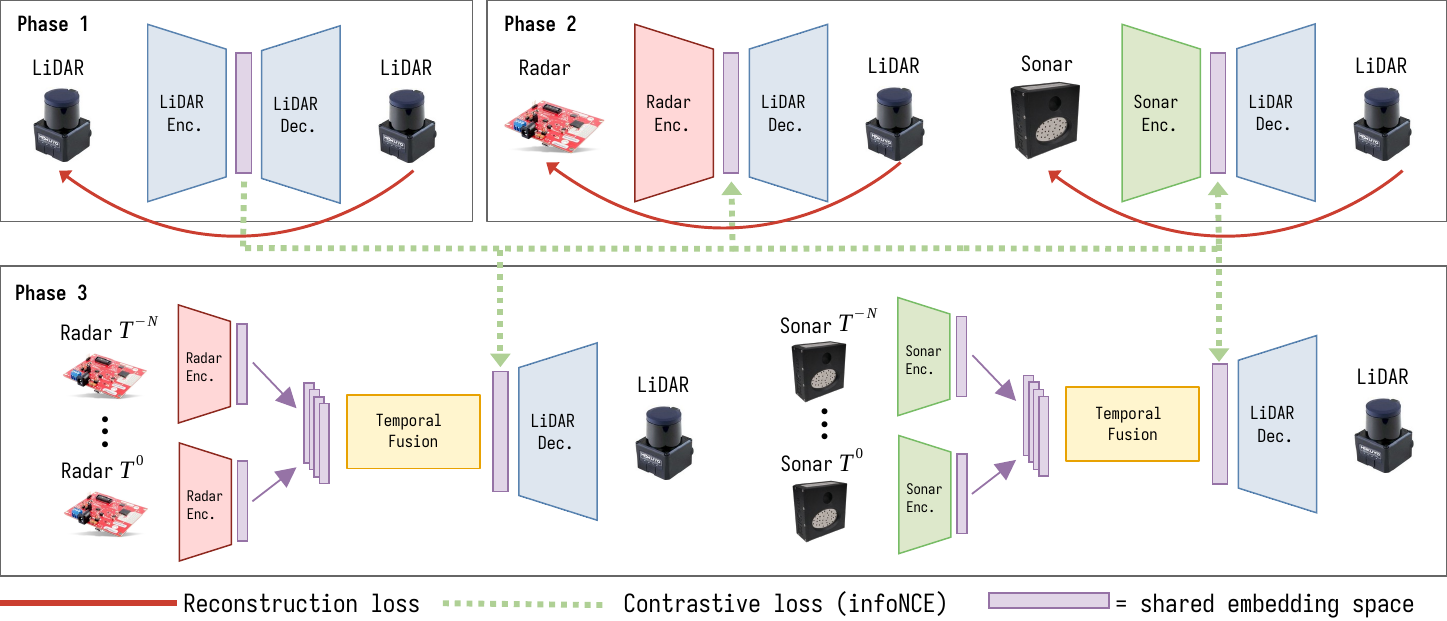}
    \caption[Updated LiDAR-BIND Training Pipeline]{Overview of the updated LiDAR-BIND training pipeline. Phase one defines the shared embeddings space, and similar to the original LiDAR-BIND framework, in phase two, we train the encoders for other modalities. In phase three, the temporal fusion module is trained to create one temporally fused embedding. Not shown here are the new losses introduced in this section to not over clutter the image. The framework now includes temporal similarity, transformation-based losses and a temporal transformer for fusing embeddings over time.}
    \label{fig:LiDARBIND_temporal_training_overview}
\end{figure*}

The proposed improvements are integrated into a modified training pipeline for the LiDAR-BIND framework, which now consists of three stages, as presented in Figure~\ref{fig:LiDARBIND_temporal_training_overview}. Each stage can be trained and deployed independently, allowing for flexibility in the training and deployment process. The stages are as follows:
\begin{enumerate}
    \item \textbf{Stage 1: LiDAR-to-LiDAR Training.} The shared embedding space is defined using LiDAR data, with the addition of the temporal similarity constraint to enforce consistency between consecutive embeddings.
    \item \textbf{Stage 2: Cross-Modality Training.} Encoders for other modalities (e.g., radar, sonar) are trained to align with the shared latent space, now also incorporating the temporal similarity loss for consecutive embeddings.
    \item \textbf{Stage 3: Temporal Transformer Training.} A temporal fusion module is trained on sequences of embeddings to produce temporally fused representations, as described above, using a combination of the contrastive loss (infoNCE), the temporal similarity and the transformation consistency loss to ensure the fused predictions maintain accurate motion dynamics.
\end{enumerate}

These improvements, both on learning objectives and architectural changes, directly address the temporal limitations identified in the original LiDAR-BIND framework, providing an approach to increase the consistency of generated frames by incorporating or fusing multiple timesteps. Our approach towards validating the effectiveness of these modifications is presented in the next section, with further discussion provided in Section~\ref{sec:LiDARBIND_temporal_discussion}.

\section{Experiments and Results}
\label{sec:LiDARBIND_temporal_experiments}
\begin{figure*}[ht]
    \centering
    \includegraphics[width=\textwidth]{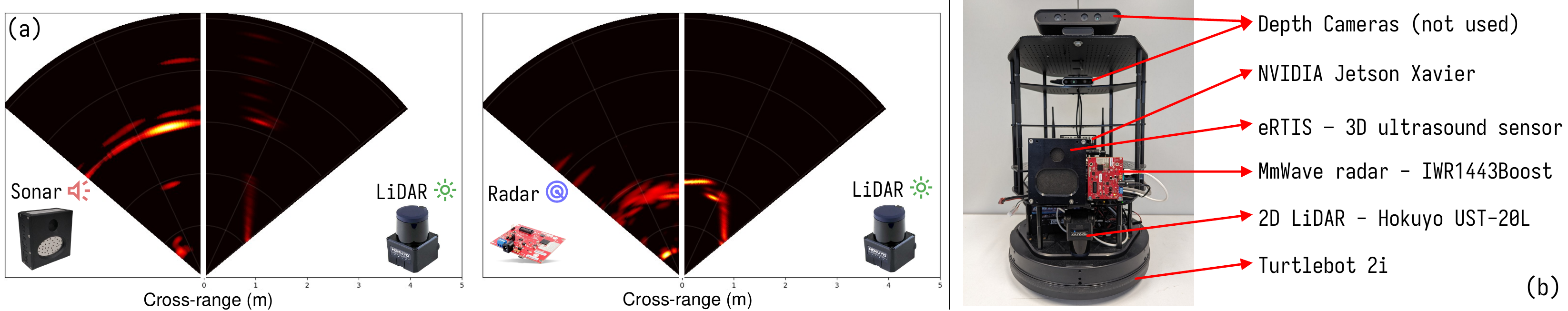}
    \caption[Robotic Platform and Sample Heatmaps]{(a) Example of processed range-azimuth heatmaps from the LiDAR-BIND dataset. These heatmaps represent processed sensor data, where for each azimuth angle, the range information is encoded in a 2D grid format. (b) The robotic platform used for data collection, equipped with a 2D LiDAR, mmWave radar, and sonar sensors.}
    \label{fig:LiDARBIND_temporal_robot}
\end{figure*}

To evaluate the proposed improvements, we focus on both the reconstruction capabilities and SLAM performance, emphasising temporal consistency and improved temporal encoding by exploring and employing several metrics. For consistency and comparability, we use the same datasets as in the prior LiDAR-BIND work.
The experimental dataset can be summarised as follows: data was collected using a small robotic platform equipped with a 2D LiDAR (Hokuyo UST-20L) \cite{hokuyo_ust-20lx_2014}, mmwave radar (TI IWR1443) \cite{texas_instruments_iwr1443_2018}, and sonar (eRTIS) \cite{Kerstens2019}, Figure \ref{fig:LiDARBIND_temporal_robot}-b depicts the used robot setup. Sensor measurements were pre-processed to select the largest common field-of-view (FOV) for all sensors, which is an azimuth field of 100 degrees and a maximum range of approximately 5 meters. Recordings took place across multiple indoor environments, including hallways, offices and residential rooms.
Our framework accommodates multiple datasets, with the mandatory condition that each contains the binding modality (LiDAR). Accordingly, the complete dataset consists of: (1) recordings in which all three sensors were used simultaneously; (2) several smaller sets covering subsets of these sensors; and (3) a simulated 2D LiDAR-only dataset employed for the initial stage of pre-training. This results in an average of 50,000 samples per sensor, and approximately 200,000 samples for LiDAR-only pre-training.
Model training follows the earlier work, splitting each dataset into training and validation sets. For contrastive learning, larger batch sizes favour learning performance; thus, we utilise a batch size of 64 (constrained by available hardware).

We use different model setups to evaluate the proposed improvements on each separate modality and a new temporal multi-modal fusion version. The previous LiDAR-BIND models are used as baseline, next to the CNN-based models that were employed as baseline in the original LiDAR-BIND work. These CNN models showed slightly better SLAM performance than the original LiDAR-BIND models. For this reason, we theorise the CNN architecture is better suited to maintain spatial information throughout the model, resulting in better spatial consistency of the measurements across time, but coming with the cost of worse encoding performance and much larger feature vectors. It is important to note that we will only evaluate decoding to LiDAR data in this paper, as the primary goal is to improve LiDAR-SLAM performance by improving the temporal consistency of predictions.

In addition to the main model comparisons, we also evaluate the individual improvements introduced in this work in the ablation study section to analyse their specific contributions to temporal consistency and SLAM performance.

The models we evaluate are as follows:
\begin{itemize}
    \item \textbf{The original LiDAR-BIND models}: R2L (radar-to-LiDAR), and S2L (sonar-to-LiDAR) models.
    \item \textbf{CNN-based baseline models}: These models use a CNN-based autoencoder architecture, inspired by U-Net \cite{ronneberger_u-net_2015}. Each model is separately trained using the same dataset as the original LiDAR-BIND work, but only using the specific modality. These models are denoted as: R2L-BASELINE, S2L-BASELINE.
    \item \textbf{The original LiDAR-BIND fusion}: This model uses the encoders of each modality and the fusion mechanism of our framework to combine multiple modalities in the embedding space. This model is denoted as Fusion; an overview of the mechanism employed in this model can be found in Section \ref{sec:LiDARBIND_temporal_lidarbind}.
    \item \textbf{LiDAR-BIND models with temporal fusion (T)}: These models combine the updated architecture and proposed improvements; temporal similarity loss, transformation loss and temporal fusion. The temporal fusion module uses five consecutive embeddings, fuses them, and enforces similarity between consecutive frames. This results in improved temporal and spatial consistency in predictions. In our tables and figures, these models are denoted as R2L-T and S2L-T, where T stands for temporal fusion. These are the separate models for each modality after training phase three, as depicted in Figure \ref{fig:LiDARBIND_temporal_training_overview}
    \item \textbf{LiDAR-BIND modality fusion with temporal transformer}: This uses the encoders of each modality and the temporal module to fuse multiple time steps, as well as the fusion mechanism of our framework to combine multiple modalities in the embedding space. This model is denoted as Temporal Fusion; an overview of the mechanism employed in this model is shown in the overview Figure \ref{fig:LiDARBIND_temporal_overview_detailed}. This version represents the improved temporal multi-modal fusion approach we propose in this work.
\end{itemize}

\begin{table*}[th]
    \centering
    \caption[Temporal-fusion Results]{Evaluation metrics for the different models are shown. We present SLAM results for two configurations: one using odometry information and one without odometry. In addition to SLAM performance, we report frame-level and scan-level metrics, as well as an adapted version of the FVMD metric as proposed by \cite{liu_frechet_2024}. Additionally, we include measurements of embedding similarity and prediction correlation peak distance. Results corresponding to the temporally modified LiDAR-BIND model are highlighted in bold.}
    \label{tab:all_metrics}
    \begin{tabular}{lcccccccccc}
    \toprule
    & \multicolumn{3}{c}{\textbf{SLAM (no odom)}} & \multicolumn{3}{c}{\textbf{SLAM (odom)}} & \textbf{EMB.} & \multicolumn{2}{c}{\textbf{TEMPORAL}} & \multicolumn{1}{c}{\textbf{FRAME}} \\ 
    \cmidrule(lr){2-4} \cmidrule(lr){5-7} \cmidrule(lr){8-8} \cmidrule(lr){9-10} \cmidrule(lr){11-11} 
 Method                  & MEAN $\downarrow$ & STD $\downarrow$   & IOU $\uparrow$   & MEAN $\downarrow$ & STD $\downarrow$  & IOU $\uparrow$  & Cos. Sim. $\uparrow$ & FVMD $\downarrow$    & Peak Dist. $\downarrow$ & PSNR $\uparrow$  \\ 
    \midrule
 LiDAR                   & 0.037 & 0.020 & -     & 0.037 & 0.020 & -     & -         & -         & -          & -      \\ 
 L2L                     & 0.057 & 0.026 & 1.000 & 0.069 & 0.033 & 1.000 & 1.000     & 920.005   & 0.693      & 45.987 \\ 
 R2L                     & 1.064 & 0.677 & 0.347 & 0.070 & 0.053 & 0.696 & 0.308     & 1737.440  & 6.021      & 37.970 \\ 
 S2L                     & 1.517 & 0.881 & 0.279 & 0.062 & 0.034 & 0.540 & 0.211     & 1854.099  & 11.642     & 37.401 \\ 
 Fusion                  & 0.077 & 0.050 & 0.584 & 0.054 & 0.034 & 0.737 & 0.725     & 1334.195  & 2.180      & 40.338 \\ 
 R2L-BASELINE            & 0.193 & 0.148 & 0.509 & 0.097 & 0.059 & 0.686 & -         & 1638.562  & 8.515      & 36.887 \\ 
 S2L-BASELINE            & 0.199 & 0.099 & 0.642 & 0.097 & 0.056 & 0.580 & -         & 1491.295  & 5.019      & 38.174 \\ 
    \midrule
    \textbf{R2L-T}          & \textbf{0.140} & 0.086 & 0.448 & 0.091 & 0.061 & 0.641 & \textbf{0.803}    & \textbf{1287.790}  & 5.230       & 38.759 \\ 
    \textbf{S2L-T}          & \textbf{0.498} & 0.321 & 0.479 & 0.086 & 0.057 & 0.659 & \textbf{0.786}    & \textbf{1521.766}  & 5.723       & 37.973 \\ 
    \textbf{Temporal Fusion}& \textbf{0.077} & 0.044 & 0.661 & 0.064 & 0.032 & 0.679 & \textbf{0.909}    & \textbf{1397.252}  & 4.767       & 41.095 \\ 
    \bottomrule
    \end{tabular}
\end{table*}

As discussed in the background section (Section~\ref{sec:LiDARBIND_temporal_metrics_sota}), metrics such as FVD and FID-VID are not well-suited to our application due to the sparse and heterogeneous nature of our sensor data. The Fréchet Video Motion Distance (FVMD) metric, proposed in~\cite{liu_frechet_2024}, is specifically designed to evaluate motion consistency in video data and is thus a more relevant and usable metric for our scenario. Unlike metrics that depend on visual frame quality, FVMD assesses the consistency of the motion of specific landmarks by analysing the velocity and acceleration of tracked points between consecutive frames. This characteristic makes FVMD particularly suitable for our application, as it aligns directly with our SLAM objectives, where scan matching or pose estimation is typically based on the relative motion between landmarks in point cloud data. To this end, we also evaluate the effectiveness of the FVMD metric for quantifying SLAM system performance.

\begin{figure*}[ht]
    \centering
    \includegraphics[width=0.90\textwidth]{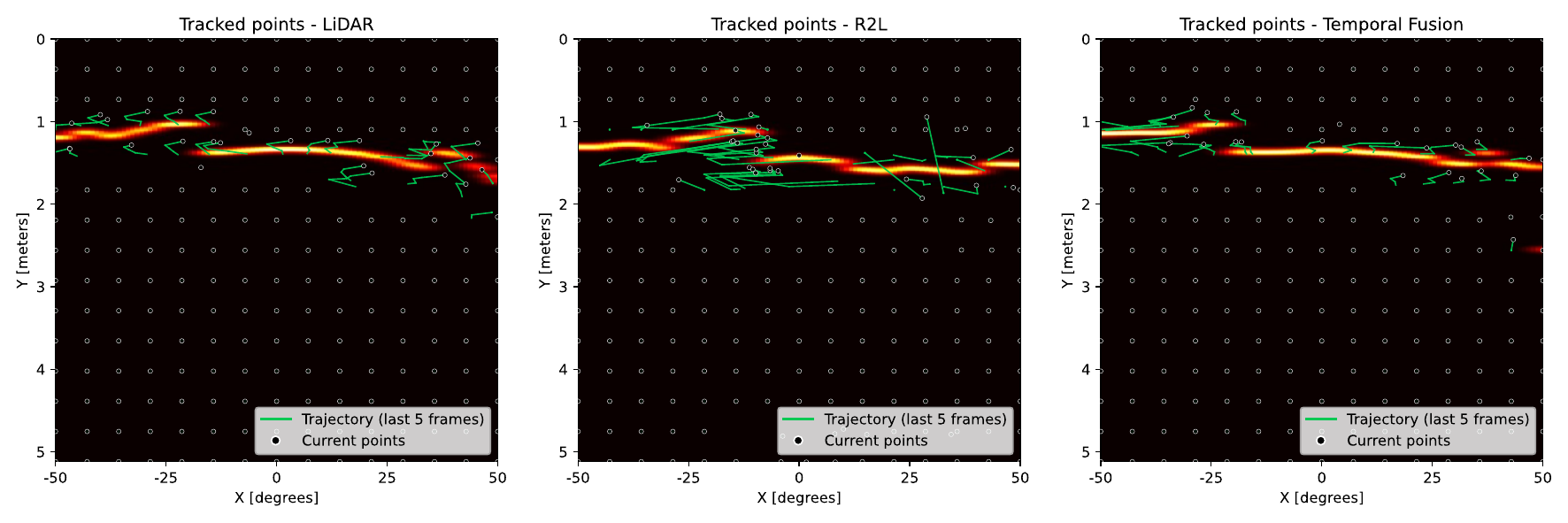}
    \caption[FVMD Lukas-Kanade Point Tracking Difference]{Overview of the flow fields obtained using the Lucas-Kanade optical flow algorithm on a grid of points for different models. The leftmost column shows the ground truth LiDAR flow field, while the other columns show the flow fields obtained from different models. The flow fields are visualised by plotting the trajectory of the tracked points. The results illustrate that the proposed improvements lead to flow fields that more closely resemble the ground truth, indicating improved temporal consistency in the predictions.}
    \label{fig:LiDARBIND_temporal_flow_field}
\end{figure*}

The FVMD score is calculated by first extracting motion trajectories of key points, typically using a pre-trained point tracking model such as PIPs++~\cite{zheng_pointodyssey_2023}. Next, the Fréchet distance between the distributions of motion trajectories in the generated and reference video sets is calculated to obtain the final FVMD score. Lower FVMD scores indicate better motion consistency in the generated data. However, since the PIPs++ model does not support our specific data modality, we instead extract motion trajectories using a simple Lucas-Kanade optical flow algorithm, applied to a pre-defined grid in our heatmap images. Figure \ref{fig:LiDARBIND_temporal_flow_field} presents example flow fields created using this method. We show the flow fields for the ground truth LiDAR data, as well as for the R2L and the temporal fusion model. These plots illustrate the differences in motion consistency between the different models, with the temporal fusion model showing a flow field more similar to the ground truth LiDAR data. Given these modifications, as well as the differences between our input data and typical usage scenarios, we cannot assume the validity of the FVMD score for our evaluation. Therefore, we further assess the metric’s effectiveness to verify its suitability for evaluating the temporal consistency of our models.

Next to this modified FVMD metric, we will also evaluate the model's predictions on different levels, which can be summarised as follows:
\begin{itemize}
    \item \textbf{Frame (heatmap) reconstruction}: We evaluate the reconstruction performance of the models by comparing the predicted heatmaps to the ground truth LiDAR heatmaps. This is done using the Peak-Signal-to-Noise-Ratio (PSNR) between the predicted and ground truth heatmaps.
    \item \textbf{Embedding similarity}: The goal of our framework is to embed the different modalities into a shared embedding space, which is defined by the LiDAR encoder. The cosine similarity between the predicted embeddings and the ground truth LiDAR embeddings is used to evaluate the performance of the models in this regard. A higher similarity (max $1$) indicates a better alignment of the predicted embeddings with the ground truth LiDAR embeddings, resulting in a better prediction with the LiDAR decoder.
    \item \textbf{SLAM performance}: We evaluate the SLAM performance of the models by using the predicted LiDAR point clouds in a SLAM pipeline, specifically Google Carthographer \cite{noauthor_cartographer_nodate}. The SLAM performance is evaluated using the Absolute Positional Error (APE), which measures the difference between the estimated trajectory and the ground truth trajectory. The ground truth trajectory is obtained from the wheel encoders of the robot.
 Furthermore, we evaluate the occupancy maps match by calculating the IoU (Intersection over Union) between an L2L map and a map generated using the predicted point clouds from another model or fusion version described above. The maps are first aligned by finding the best transformation between the two maps using the cross-correlation between the two maps. We chose the L2L map as a reference here, as this map contains the amount of detail we can expect the SLAM algorithm to produce based on the model predictions. We show the SLAM performance metrics for two SLAM configurations: one where scan matching and odometry are used, and another where scan matching is the only source of transformation between the scans to obtain pose information.
    \item \textbf{Temporal metrics}: We evaluate the temporal consistency of the models using the FVMD metric, as described above. This metric measures the consistency of motion patterns in the predicted data, which is crucial for SLAM performance. We also calculate the 2D cross-correlation between consecutive predictions and compare this to the ground truth LiDAR cross-correlation. The distance between the peaks in these correlation maps is used as a measure of how well the predicted motion matches the ground truth motion. A smaller distance indicates a better match.
\end{itemize}

\begin{figure*}[ht]
    \centering
    \includegraphics[width=0.95\textwidth]{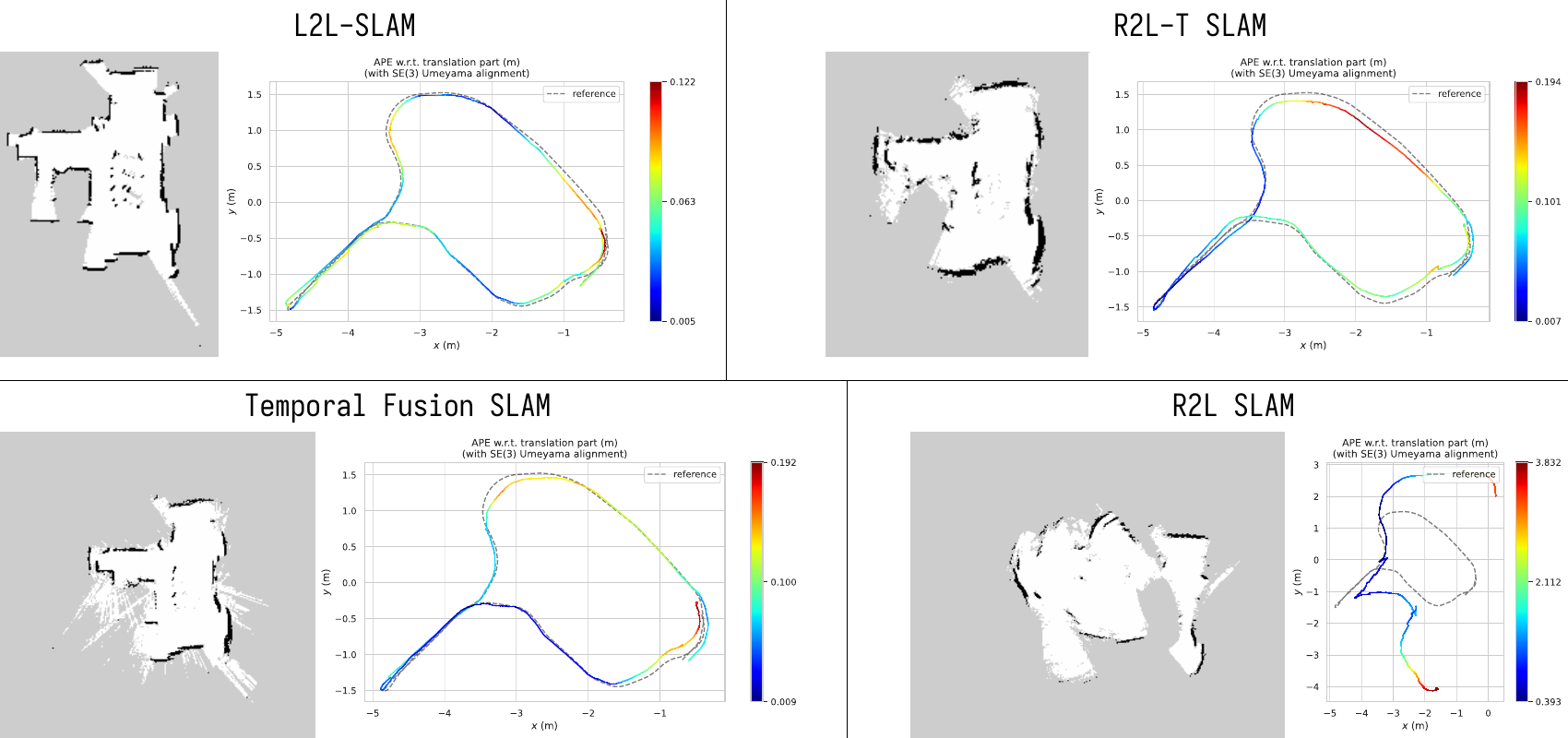}
    \caption[LiDARBIND Temporal SLAM Results Overview]{SLAM results for the different configurations are presented. In all cases, the SLAM system has been configured to rely primarily on scan matching to determine transformations between sequential scans. We provide occupancy maps and pose graphs for the L2L configuration (as a reference), as well as for the R2L-T model, the original R2L model, and the Temporal Fusion SLAM method. The results demonstrate that our proposed modifications lead to a considerable improvement in SLAM performance, particularly when compared to R2L SLAM, which utilises the original LiDAR-BIND model. The occupancy maps generated by the modified models are notably more consistent, and the pose graphs show improved alignment with the ground truth trajectory.}
    \label{fig:LiDARBIND_temporal_slam_results}
\end{figure*}

The results of this evaluation are presented in Table \ref{tab:all_metrics}. The table shows the specified metrics for each model across our evaluation datasets, which consist of synchronised LiDAR, sonar and mmWave radar measurements. Figure \ref{fig:LiDARBIND_temporal_slam_results} presents the occupancy grids and pose-graphs for several radar-based prediction models, visualising the SLAM results for the case where no odometry is used, and thus the reliance on scan matching and consistent predictions is highlighted.

We observe that the improvements made to the LiDAR-BIND framework result in increased SLAM performance, particularly in situations where scan matching serves as the primary mechanism for determining transformations between consecutive scans, indicating improved consistency between generated frames. The SLAM metrics from Table~\ref{tab:all_metrics} confirm this observation and indicate a substantial reduction in absolute positional error (APE) for the R2L-T model when compared to the original R2L model. Furthermore, there is an increase in the intersection over union (IoU) score of the corresponding occupancy maps. Similar trends are observed for the S2L-T model, which shows improvement over the original S2L model in terms of SLAM performance.

Nevertheless, the S2L-T SLAM model does not perform as well as the comparable radar-based approach. The precise reason for this discrepancy remains unclear. In this case, the performance of the S2L-BASELINE model is still superior, which illustrates the ongoing trade-off between the modularity and flexibility provided by the LiDAR-BIND framework and the enhanced performance associated with a dedicated latent representation space. It is possible that the encoding approach we use for each modality can be specifically adapted to the modality to further increase performance; this is out of scope for this work, focusing on increasing the temporal consistency of predictions, which the results still indicate the improved method succeeds in doing.

Next, the temporal transformer fusion model also demonstrates significant improvements in SLAM results, including lower APE and higher IoU, relative to the original LiDAR-BIND fusion model. The occupancy maps produced by the improved models are more consistent and exhibit greater alignment with the ground truth trajectory, as illustrated in Figure~\ref{fig:LiDARBIND_temporal_slam_results}.
We also observe higher cosine similarity across the improved model versions compared to the original LiDAR-BIND models, indicating that the embeddings of the predicted frames are more similar to those produced by the L2L embeddings. This increased similarity should result in improved decoder performance, as embeddings are used as input to this module. The better aligned the predicted embeddings are with the L2L embeddings, the closer the predictions should match their LiDAR counterparts. The L2L model continues to offer the best overall performance across all metrics.

Finally, we observe that our metrics for evaluating temporal consistency, namely FVMD and the peak distance in the cross-correlation maps, provide evidence that the proposed improvements lead to lower scores for these metrics, as expected. The reduced FVMD scores in the improved models indicate greater consistency in the motion patterns of the predicted data. Similarly, the smaller peak distances in the cross-correlation maps suggest that the predicted motion aligns better with the ground truth motion. This increased consistency is particularly important for SLAM performance, as it ensures that the predicted frames are coherent both with each other and with the ground truth trajectory.

To further assess the validity of these metrics, we calculate the Spearman’s rank correlation, Pearson’s correlation and Kendall’s Tau correlation coefficients between these metrics and the mean absolute positional error (APE) for the SLAM configuration using no odometry as input. For the FVMD metric, we found a strong and statistically significant correlation across the presented results (Spearman’s $\rho$ = 0.88, p = 0.0016, Kendall $\tau$ = 0.72, p = 0.0059) and a substantial linear relationship (Pearson’s r = 0.76, p = 0.017). For the 2D correlation map peak distance, there was also a strong and statistically significant correlation across models (Spearman’s $\rho$ = 0.85, p = 0.003, Kendall $\tau$ = 0.72, p = 0.0059), as well as a substantial linear relationship (Pearson’s r = 0.73, p = 0.026). These results indicate that both metrics can be effective indicators for evaluating the SLAM performance of the models, supporting the conclusion that the proposed improvements contribute to increased temporal consistency. However, as this analysis is based on the data presented in Table~\ref{tab:all_metrics}, the sample size is limited. Therefore, these results should be interpreted with caution.

\begin{table*}[ht]
    \centering
    \caption[Impact of Temporal Window Size]{Ablation study on the impact of temporal window size. The results show the performance of the temporal fusion and temporal windowing models with different window sizes on relevant evaluation metrics. Results are obtained using different configurations of the temporarily improved R2L model (with and without temporal fusion) using a SLAM configuration to emphasise scan matching.}
    \label{tab:ablation_window_size}
    \begin{tabular}{lccccc}
    \toprule
    \textbf{Window Size and Fusion} & \textbf{APE Mean} $\downarrow$ & \textbf{APE STD.} $\downarrow$ & \textbf{Cos. Sim.} $\uparrow$ & \textbf{FVMD} $\downarrow$ \\
    \midrule
 Temporal Fusion (3) & 0.134 & 0.109 & 0.797 & 1579.900 \\
    \textbf{Temporal Fusion (5)} & \textbf{0.095} & \textbf{0.054} & \textbf{0.803} & \textbf{1287.790} \\
 Temporal Fusion (7) & 0.305 & 0.248 & 0.713 & 3346.542 \\
    \midrule
 Windowing (3) & 0.158 & 0.081 & 0.587 & 1522.971 \\
 Windowing (5) & 0.154 & 0.099 & 0.589 & 1475.530 \\
 Windowing (7) & 0.218 & 0.143 & 0.590 & 1614.833 \\
    \bottomrule
    \end{tabular}
\end{table*}

\section{Ablation study}
\label{sec:LiDARBIND_temporal_ablation}
In this section, we present an ablation study to evaluate the contribution of specific components in our proposed enhancements to the LiDAR-BIND framework. In particular, we investigate the influence of the temporal window size and compare the two strategies for temporal integration: (1) temporal windowing, which aggregates information by averaging over multiple time steps, and (2) temporal fusion, in which a CNN-based module is employed to fuse embeddings from several consecutive time steps. Next, we investigate the impact of contrastive loss in the second stage compared to a cosine similarity as loss function. We also evaluate at which stage measurement fusion could be applied. We present fusion after the encoder (using the proposed temporal fusion) compared to combining multiple time steps by creating an encoder that takes multiple time steps as input. Finally, we study the effect of the improved architecture by applying the proposed training objectives (phase two) on the old LiDAR-BIND architecture.

\subsection{Impact of Window Size and Temporal Windowing vs. Temporal Fusion}
To evaluate the impact of window size on SLAM performance, we trained fusion models with different window sizes: 3, 5, and 7. The SLAM results are presented in Table~\ref{tab:ablation_window_size} for radar-based models using a SLAM configuration where odometry and scan matching are used for pose estimation, where the scan matching pose is preferred by configured weights. For comparison, we also include the results obtained using the windowing approach based on averaging the embeddings over multiple time steps. We change here to a weighted configuration, as this windowing approach consistently failed to generate a valid pose in the case with no odometry as input. This way, we can compare the two strategies more effectively, and we show that improved temporal quality of the predictions can also benefit SLAM performance when odometry is available.

We find that, as expected, temporal fusion performs better than simple averaging. In particular, there is a significant improvement in cosine similarity between the fused embeddings and the corresponding LiDAR embedding for each timestep. SLAM performance is also generally higher for temporal fusion, except at the largest window sizes. Additionally, we observe that increasing the window size slightly improves model performance up to a point. However, extending the window size further does not continue to yield improvements. This plateau or decline in performance may be due to the increased delay introduced when averaging over more time steps, or to the accumulation of unrelated information, which acts as noise that the temporal transformer must then handle as the window expands.

\subsection{Early fusion vs. latent fusion}
To compare early fusion and latent temporal fusion, we made minor adaptations to our encoder architecture to enable it to process multiple time steps as input. Specifically, the encoder was adapted to receive and process a sequence of time steps simultaneously, rather than handling one time step at a time followed by fusion at the embedding level. This modified architecture employs 3D convolutional layers to effectively process the temporal dimension alongside the multiple range-azimuth heatmaps. We compare this early fusion model to the latent fusion model, both with a window of 5 samples.

As shown in Table~\ref{tab:ablation_fusion_method}, the early fusion approach achieves performance comparable to that of the latent fusion approach. This result can be attributed to the capacity of the modified encoder architecture to directly learn temporal relationships across the additional temporal dimension during encoding. In contrast, the latent fusion approach uses a temporal transformer to capture these temporal relations at the more abstract embedding level. We observe that the early fusion method achieves an overall higher embedding cosine similarity, which suggests that the multi-input encoder can align the embeddings more closely with those from the phase one L2L model. On the other hand, the superior SLAM performance of the latent fusion method can be attributed to the ability of the temporal fusion module to smooth embeddings over time, reducing sudden changes in the predictions, and consequently improving the SLAM results.

\begin{table}[ht]
    \centering
    \caption[Early Fusion vs. Latent Fusion]{Ablation study on early fusion vs. latent fusion. Results are obtained using the temporally improved R2L model and a modified version of the R2L model to use more input measurements, using a SLAM configuration to emphasise scan matching.}
    \label{tab:ablation_fusion_method}
    \begin{tabular}{lccccc}
    \toprule
    \textbf{Fusion stage} & \textbf{APE Mean} $\downarrow$ & \textbf{STD.} $\downarrow$ & \textbf{Cos. Sim.} $\uparrow$ & \textbf{FVMD} $\downarrow$ \\
    \midrule
 Early Fusion        & 0.120 & 0.072 & 0.892 & 1494.904 \\
 Latent Fusion       & 0.095 & 0.054 & 0.803 & 1287.790 \\
    \bottomrule
    \end{tabular}
\end{table}

\begin{table*}[ht]
    \centering
    \caption[Embedding Loss Functions Ablation Study]{Ablation study on embedding loss functions. We show SLAM performance and embedding similarity for contrastive loss variants and temporal similarity loss. Results are from R2L models after second phase training, using a SLAM configuration to emphasise scan matching.}
    \label{tab:ablation_embedding_loss}
    \begin{tabular}{lccccc}
    \toprule
    \textbf{Embedding loss} & \textbf{APE Mean} $\downarrow$ & \textbf{APE Mean STD.} $\downarrow$ & \textbf{Cos. Sim.} $\uparrow$ & \textbf{FVMD} $\downarrow$ \\
    \midrule
 InfoNCE                         & 1.026 & 0.486 & 0.814 & 1751.830 \\
 InfoNCE + Temporal Similarity   & 0.877 & 0.436 & 0.819 & 1569.666 \\
 InfoNCE modified                & 1.370 & 0.712 & 0.840 & 1719.935 \\
    \bottomrule
    \end{tabular}
\end{table*}

\subsection{Impact of Cosine Similarity vs. Contrastive Loss}
As explained in Section~\ref{sec:LiDARBIND_temporal_method}, in the second training phase of our approach, we employ a temporal similarity loss, specifically implemented as a cosine similarity function, to strictly enforce similarity between consecutive samples. During this phase, we also use a contrastive loss function (the InfoNCE loss) to align cross-modal embeddings. We identified a potential conflict in the training objective, as the temporal samples we are aiming to align may also be considered negative samples by the contrastive loss. This conflict becomes more likely as the batch size increases. Nevertheless, we believe that explicitly training for consistency between temporal samples yields greater prediction consistency and improved SLAM performance. With our chosen batch size of 64, the likelihood of such conflicts in the training objective is very low. In this ablation study, we investigate the difference made by adding the temporal similarity loss and evaluate its impact on overall embedding similarity and SLAM performance.

We also conducted experiments using a variant of the InfoNCE loss incorporating a regularisation term that reduces the penalty for temporally or spatially close samples being treated as negative pairs. Similar to the cosine similarity, this modification aims to improve alignment between temporally close samples. However, this approach for encouraging temporal embedding consistency is less direct than the explicit temporal similarity loss introduced above. We present the results of both modifications in this section.

The results, presented in Table~\ref{tab:ablation_embedding_loss}, show that while the modified infoNCE object obtains greater embedding similarity, its SLAM performance is still degraded, likely because temporal inconsistency is still more likely to occur. The InfoNCE loss with temporal similarity, however, achieves a better balance between embedding similarity and SLAM performance. This indicates that the explicit temporal similarity loss effectively enforces consistency between consecutive embeddings, leading to improved SLAM results. The original InfoNCE loss without temporal similarity performs worse than the modified InfoNCE loss, indicating that the temporal consistency is indeed beneficial for SLAM performance.

\subsection{Optimised Architecture for Spatial Encoding}
As a final ablation, we evaluate the impact of the changes we made to the encoder and decoder architectures, as described in Section~\ref{sec:LiDARBIND_temporal_method}. We apply the proposed training objectives for phase two on the old LiDAR-BIND architecture and compare the results to those obtained using the updated architecture. The results, presented in Table~\ref{tab:ablation_architecture}, show that the updated architecture achieves better SLAM performance and embedding similarity compared to the original architecture. This indicates that the architectural changes we made are effective in improving the model's ability to encode spatial information, leading to better alignment of embeddings and improved SLAM results.
This test also confirms our proposed training objective for forcing embedding similarity between time-steps is effective at improving the temporal quality of the predictions.

This result also aligns with our intuition that the transformer and fully-connected layers in the original architecture were not fully leveraging and maintaining the spatial information present in the LiDAR data. By optimising the architecture for spatial encoding and relying more on convolutional layers, which are better suited for capturing local patterns, we can better capture the spatial relationships, leading to improved temporal consistency and SLAM performance.

\begin{table*}[ht]
    \centering
    \caption[Embedding Loss Functions Ablation Study]{Ablation study on embedding loss functions. We show SLAM performance and embedding similarity for the different architectures and training strategies. Results are from R2L models after second phase training, using a SLAM configuration to emphasise scan matching.}
    \label{tab:ablation_architecture}
    \begin{tabular}{lccccc}
    \toprule
    \textbf{Embedding loss} & \textbf{APE Mean} $\downarrow$ & \textbf{APE Mean STD.} $\downarrow$ & \textbf{Cos. Sim.} $\uparrow$ & \textbf{FVMD} $\downarrow$ \\
    \midrule
 Original R2L                                & 1.064 & 0.677 & 0.308 & 1737.440 \\
 Original R2L with proposed improvements     & 0.926 & 0.511 & 0.574 & 1704.944 \\
 New R2L                                     & 0.877 & 0.436 & 0.819 & 1569.666 \\
    \bottomrule
    \end{tabular}
\end{table*}

\section{Discussion and Conclusion}
\label{sec:LiDARBIND_temporal_discussion}
In this work, we improved the spatial and temporal consistency of predictions in the LiDAR-BIND multi-modal sensor fusion framework. We diagnosed the original framework’s limitation in temporal coherence for downstream SLAM and, inspired by state-of-the-art temporal modelling, introduced architectural and training changes together with domain-appropriate temporal metrics. Overall, our contributions are:
\begin{itemize}
    \item Modular temporal fusion that improves temporal consistency and SLAM while preserving LiDAR-BIND’s plug-and-play modularity.
    \item Temporal and motion consistency training objectives and architectural updates that enforce coherent motion and preserve spatial structure.
    \item Domain-suitable temporal metrics (adapted FVMD and correlation-peak distance) that correlate with SLAM accuracy.
    \item Empirical gains and clarified trade-off: our methods narrow the gap between single-modality optimised latents, presented in section \ref{sec:LiDARBIND_temporal_lidarbind}, and the shared latent while retaining modular, adaptive multi-modal fusion.
\end{itemize}

The proposed method remains fully modular: at runtime, each perception modality can be used independently or fused adaptively, and the temporal fusion module can be trained and deployed independently. This preserves LiDAR-BIND’s ability to add new modalities post-deployment while improving temporal stability. Our results explicitly quantify the trade-off between optimising the latent for a single modality—yielding higher temporal consistency and SLAM accuracy—and learning a shared latent that enables modularity and adaptive fusion at some cost to accuracy and temporal consistency. The proposed temporal training and fusion strategies close much of this gap while keeping the shared latent and its practical benefits.

Future work includes exploring the updated framework's capabilities for single modality anomaly detection by tracking embedding orientation across time. Next to that, exploring attention-based and recurrent temporal modules for longer-range dependencies, and refining modality-specific encoders to further improve temporal stability and SLAM across diverse environments. Finally, exploring dynamic windowing could be a promising strategy to deal with different sensor update rates and motion dynamics.

\bibliographystyle{IEEEtran}
\bibliography{export}

\begin{thebibliography}{10}
\providecommand{\url}[1]{#1}
\csname url@samestyle\endcsname
\providecommand{\newblock}{\relax}
\providecommand{\bibinfo}[2]{#2}
\providecommand{\BIBentrySTDinterwordspacing}{\spaceskip=0pt\relax}
\providecommand{\BIBentryALTinterwordstretchfactor}{4}
\providecommand{\BIBentryALTinterwordspacing}{\spaceskip=\fontdimen2\font plus
\BIBentryALTinterwordstretchfactor\fontdimen3\font minus \fontdimen4\font\relax}
\providecommand{\BIBforeignlanguage}[2]{{%
\expandafter\ifx\csname l@#1\endcsname\relax
\typeout{** WARNING: IEEEtran.bst: No hyphenation pattern has been}%
\typeout{** loaded for the language `#1'. Using the pattern for}%
\typeout{** the default language instead.}%
\else
\language=\csname l@#1\endcsname
\fi
#2}}
\providecommand{\BIBdecl}{\relax}
\BIBdecl

\bibitem{wang_survey_2025}
\BIBentryALTinterwordspacing
H.~Wang, J.~Liu, H.~Dong, and Z.~Shao, ``\BIBforeignlanguage{en}{A {Survey} of the {Multi}-{Sensor} {Fusion} {Object} {Detection} {Task} in {Autonomous} {Driving}},'' \emph{\BIBforeignlanguage{en}{Sensors}}, vol.~25, no.~9, p. 2794, Jan. 2025, publisher: Multidisciplinary Digital Publishing Institute. [Online]. Available: \url{https://www.mdpi.com/1424-8220/25/9/2794}
\BIBentrySTDinterwordspacing

\bibitem{li_multi-sensor_2025}
\BIBentryALTinterwordspacing
W.~Li, X.~Wan, Z.~Ma, and Y.~Hu, ``\BIBforeignlanguage{en}{Multi-sensor {Fusion} {Perception} of {Vehicle} {Environment} and its {Application} in {Obstacle} {Avoidance} of {Autonomous} {Vehicle}},'' \emph{\BIBforeignlanguage{en}{International Journal of Intelligent Transportation Systems Research}}, vol.~23, no.~1, pp. 450--463, Apr. 2025. [Online]. Available: \url{https://doi.org/10.1007/s13177-024-00460-x}
\BIBentrySTDinterwordspacing

\bibitem{zhang_perception_2023}
\BIBentryALTinterwordspacing
Y.~Zhang, A.~Carballo, H.~Yang, and K.~Takeda, ``Perception and sensing for autonomous vehicles under adverse weather conditions: {A} survey,'' \emph{ISPRS Journal of Photogrammetry and Remote Sensing}, vol. 196, pp. 146--177, Feb. 2023. [Online]. Available: \url{https://www.sciencedirect.com/science/article/pii/S0924271622003367}
\BIBentrySTDinterwordspacing

\bibitem{balemans_lidar-bind_2024}
\BIBentryALTinterwordspacing
N.~Balemans, A.~Anwar, J.~Steckel, and S.~Mercelis, ``{LiDAR}-{BIND}: {Multi}-{Modal} {Sensor} {Fusion} {Through} {Shared} {Latent} {Embeddings},'' \emph{IEEE Robotics and Automation Letters}, pp. 1--8, 2024, conference Name: IEEE Robotics and Automation Letters. [Online]. Available: \url{https://ieeexplore.ieee.org/document/10670298}
\BIBentrySTDinterwordspacing

\bibitem{balemans_r2l-slam_2023}
\BIBentryALTinterwordspacing
N.~Balemans, L.~Hooft, P.~Reiter, A.~Anwar, J.~Steckel, and S.~Mercelis, ``{R2L}-{SLAM}: {Sensor} {Fusion}-{Driven} {SLAM} {Using} {mmWave} {Radar}, {LiDAR} and {Deep} {Neural} {Networks},'' in \emph{2023 {IEEE} {SENSORS}}, Oct. 2023, pp. 1--4, iSSN: 2168-9229. [Online]. Available: \url{https://ieeexplore.ieee.org/abstract/document/10324990}
\BIBentrySTDinterwordspacing

\bibitem{balemans_s2l-slam_2021}
\BIBentryALTinterwordspacing
N.~Balemans, P.~Hellinckx, S.~Latré, P.~Reiter, and J.~Steckel, ``{S2L}-{SLAM}: {Sensor} {Fusion} {Driven} {SLAM} using {Sonar}, {LiDAR} and {Deep} {Neural} {Networks},'' in \emph{2021 {IEEE} {Sensors}}, Oct. 2021, pp. 1--4, iSSN: 2168-9229. [Online]. Available: \url{https://ieeexplore.ieee.org/document/9639772}
\BIBentrySTDinterwordspacing

\bibitem{Kerstens2019}
\BIBentryALTinterwordspacing
R.~Kerstens, D.~Laurijssen, and J.~Steckel, ``{eRTIS}: {A} {Fully} {Embedded} {Real} {Time} {3D} {Imaging} {Sonar} {Sensor} for {Robotic} {Applications},'' in \emph{2019 {International} {Conference} on {Robotics} and {Automation} ({ICRA})}.\hskip 1em plus 0.5em minus 0.4em\relax IEEE, May 2019, pp. 1438--1443. [Online]. Available: \url{https://ieeexplore.ieee.org/document/8794419/}
\BIBentrySTDinterwordspacing

\bibitem{liu_frechet_2024}
\BIBentryALTinterwordspacing
J.~Liu, Y.~Qu, Q.~Yan, X.~Zeng, L.~Wang, and R.~Liao, ``Fréchet {Video} {Motion} {Distance}: {A} {Metric} for {Evaluating} {Motion} {Consistency} in {Videos},'' Jul. 2024, arXiv:2407.16124. [Online]. Available: \url{http://arxiv.org/abs/2407.16124}
\BIBentrySTDinterwordspacing

\bibitem{chen_video_2025}
\BIBentryALTinterwordspacing
S.~Chen, H.~Guo, S.~Zhu, F.~Zhang, Z.~Huang, J.~Feng, and B.~Kang, ``Video {Depth} {Anything}: {Consistent} {Depth} {Estimation} for {Super}-{Long} {Videos},'' Jan. 2025, arXiv:2501.12375 [cs]. [Online]. Available: \url{http://arxiv.org/abs/2501.12375}
\BIBentrySTDinterwordspacing

\bibitem{chakrabarti_temporally-consistent_2025}
\BIBentryALTinterwordspacing
A.~Chakrabarti, I.~Nayak, and D.~Goswami, ``Temporally-{Consistent} {Bilinearly} {Recurrent} {Autoencoders} for {Control} {Systems},'' Mar. 2025, arXiv:2503.19085 [eess] version: 1. [Online]. Available: \url{http://arxiv.org/abs/2503.19085}
\BIBentrySTDinterwordspacing

\bibitem{somma_hybrid_2025}
\BIBentryALTinterwordspacing
M.~Somma, ``\BIBforeignlanguage{en}{Hybrid {Temporal} {Differential} {Consistency} {Autoencoder} for {Efficient} and {Sustainable} {Anomaly} {Detection} in {Cyber}-{Physical} {Systems}},'' Apr. 2025, arXiv:2504.06320 [cs]. [Online]. Available: \url{http://arxiv.org/abs/2504.06320}
\BIBentrySTDinterwordspacing

\bibitem{wei_t-mae_2024}
\BIBentryALTinterwordspacing
W.~Wei, F.~K. Nejadasl, T.~Gevers, and M.~R. Oswald, ``T-{MAE}: {Temporal} {Masked} {Autoencoders} for {Point} {Cloud} {Representation} {Learning},'' Jul. 2024, arXiv:2312.10217 [cs] version: 3. [Online]. Available: \url{http://arxiv.org/abs/2312.10217}
\BIBentrySTDinterwordspacing

\bibitem{liang_movideo_2024}
\BIBentryALTinterwordspacing
J.~Liang, Y.~Fan, K.~Zhang, R.~Timofte, L.~V. Gool, and R.~Ranjan, ``{MoVideo}: {Motion}-{Aware} {Video} {Generation} with {Diffusion} {Models},'' Jul. 2024, arXiv:2311.11325 [cs]. [Online]. Available: \url{http://arxiv.org/abs/2311.11325}
\BIBentrySTDinterwordspacing

\bibitem{ren_consisti2v_2024}
\BIBentryALTinterwordspacing
W.~Ren, H.~Yang, G.~Zhang, C.~Wei, X.~Du, W.~Huang, and W.~Chen, ``{ConsistI2V}: {Enhancing} {Visual} {Consistency} for {Image}-to-{Video} {Generation},'' Jul. 2024, arXiv:2402.04324 [cs]. [Online]. Available: \url{http://arxiv.org/abs/2402.04324}
\BIBentrySTDinterwordspacing

\bibitem{ma_latte_2025}
\BIBentryALTinterwordspacing
X.~Ma, Y.~Wang, X.~Chen, G.~Jia, Z.~Liu, Y.-F. Li, C.~Chen, and Y.~Qiao, ``Latte: {Latent} {Diffusion} {Transformer} for {Video} {Generation},'' May 2025, arXiv:2401.03048 [cs]. [Online]. Available: \url{http://arxiv.org/abs/2401.03048}
\BIBentrySTDinterwordspacing

\bibitem{zhang_training-free_2025}
\BIBentryALTinterwordspacing
X.~Zhang, Z.~Duan, D.~Gong, and L.~Liu, ``\BIBforeignlanguage{en}{Training-{Free} {Motion}-{Guided} {Video} {Generation} with {Enhanced} {Temporal} {Consistency} {Using} {Motion} {Consistency} {Loss}},'' Jan. 2025, arXiv:2501.07563 [cs]. [Online]. Available: \url{http://arxiv.org/abs/2501.07563}
\BIBentrySTDinterwordspacing

\bibitem{yan_temporally_2023}
\BIBentryALTinterwordspacing
W.~Yan, D.~Hafner, S.~James, and P.~Abbeel, ``\BIBforeignlanguage{en}{Temporally {Consistent} {Transformers} for {Video} {Generation}},'' in \emph{\BIBforeignlanguage{en}{Proceedings of the 40th {International} {Conference} on {Machine} {Learning}}}.\hskip 1em plus 0.5em minus 0.4em\relax PMLR, Jul. 2023, pp. 39\,062--39\,098, iSSN: 2640-3498. [Online]. Available: \url{https://proceedings.mlr.press/v202/yan23b.html}
\BIBentrySTDinterwordspacing

\bibitem{wang_videolcm_2023}
\BIBentryALTinterwordspacing
X.~Wang, S.~Zhang, H.~Zhang, Y.~Liu, Y.~Zhang, C.~Gao, and N.~Sang, ``{VideoLCM}: {Video} {Latent} {Consistency} {Model},'' Dec. 2023, arXiv:2312.09109 [cs]. [Online]. Available: \url{http://arxiv.org/abs/2312.09109}
\BIBentrySTDinterwordspacing

\bibitem{unterthiner_towards_2019}
\BIBentryALTinterwordspacing
T.~Unterthiner, S.~v. Steenkiste, K.~Kurach, R.~Marinier, M.~Michalski, and S.~Gelly, ``Towards {Accurate} {Generative} {Models} of {Video}: {A} {New} {Metric} \& {Challenges},'' Mar. 2019, arXiv:1812.01717 [cs]. [Online]. Available: \url{http://arxiv.org/abs/1812.01717}
\BIBentrySTDinterwordspacing

\bibitem{heusel_gans_2018}
\BIBentryALTinterwordspacing
M.~Heusel, H.~Ramsauer, T.~Unterthiner, B.~Nessler, and S.~Hochreiter, ``{GANs} {Trained} by a {Two} {Time}-{Scale} {Update} {Rule} {Converge} to a {Local} {Nash} {Equilibrium},'' Jan. 2018, arXiv:1706.08500 [cs]. [Online]. Available: \url{http://arxiv.org/abs/1706.08500}
\BIBentrySTDinterwordspacing

\bibitem{huang_vbench_2023}
\BIBentryALTinterwordspacing
Z.~Huang, Y.~He, J.~Yu, F.~Zhang, C.~Si, Y.~Jiang, Y.~Zhang, T.~Wu, Q.~Jin, N.~Chanpaisit, Y.~Wang, X.~Chen, L.~Wang, D.~Lin, Y.~Qiao, and Z.~Liu, ``{VBench}: {Comprehensive} {Benchmark} {Suite} for {Video} {Generative} {Models},'' Nov. 2023, arXiv:2311.17982. [Online]. Available: \url{http://arxiv.org/abs/2311.17982}
\BIBentrySTDinterwordspacing

\bibitem{huang_vbench_2024}
\BIBentryALTinterwordspacing
Z.~Huang, F.~Zhang, X.~Xu, Y.~He, J.~Yu, Z.~Dong, Q.~Ma, N.~Chanpaisit, C.~Si, Y.~Jiang, Y.~Wang, X.~Chen, Y.-C. Chen, L.~Wang, D.~Lin, Y.~Qiao, and Z.~Liu, ``{VBench}++: {Comprehensive} and {Versatile} {Benchmark} {Suite} for {Video} {Generative} {Models},'' Nov. 2024, arXiv:2411.13503. [Online]. Available: \url{http://arxiv.org/abs/2411.13503}
\BIBentrySTDinterwordspacing

\bibitem{zheng_pointodyssey_2023}
\BIBentryALTinterwordspacing
Y.~Zheng, A.~W. Harley, B.~Shen, G.~Wetzstein, and L.~J. Guibas, ``{PointOdyssey}: {A} {Large}-{Scale} {Synthetic} {Dataset} for {Long}-{Term} {Point} {Tracking},'' Jul. 2023, arXiv:2307.15055 [cs]. [Online]. Available: \url{http://arxiv.org/abs/2307.15055}
\BIBentrySTDinterwordspacing

\bibitem{Oord2018}
\BIBentryALTinterwordspacing
A.~v.~d. Oord, Y.~Li, and O.~Vinyals, ``Representation {Learning} with {Contrastive} {Predictive} {Coding},'' Jul. 2018, arXiv: 1807.03748. [Online]. Available: \url{http://arxiv.org/abs/1807.03748}
\BIBentrySTDinterwordspacing

\bibitem{balemans_multi-modal_2024}
\BIBentryALTinterwordspacing
N.~Balemans, A.~Anwar, J.~Steckel, and S.~Mercelis, ``Multi-{Modal} {Sensor} {Fusion} in {Latent} {Embedding} {Space} for {Robust} {Autonomous} {Navigation},'' in \emph{2024 {IEEE} {SENSORS}}, Oct. 2024, pp. 1--4, iSSN: 2168-9229. [Online]. Available: \url{https://ieeexplore.ieee.org/document/10785240}
\BIBentrySTDinterwordspacing

\bibitem{Dosovitskiy2020}
\BIBentryALTinterwordspacing
A.~Dosovitskiy, L.~Beyer, A.~Kolesnikov, D.~Weissenborn, X.~Zhai, T.~Unterthiner, M.~Dehghani, M.~Minderer, G.~Heigold, S.~Gelly, J.~Uszkoreit, and N.~Houlsby, ``An {Image} is {Worth} 16x16 {Words}: {Transformers} for {Image} {Recognition} at {Scale},'' Oct. 2020, arXiv: 2010.11929. [Online]. Available: \url{http://arxiv.org/abs/2010.11929}
\BIBentrySTDinterwordspacing

\bibitem{noauthor_cartographer_nodate}
\BIBentryALTinterwordspacing
``Cartographer — {Cartographer} documentation.'' [Online]. Available: \url{https://google-cartographer.readthedocs.io/en/latest/}
\BIBentrySTDinterwordspacing

\bibitem{hokuyo_ust-20lx_2014}
{Hokuyo}, ``{UST}-{20LX} ({UUST004}) {Specification},'' 2014, pages: 4-9.

\bibitem{texas_instruments_iwr1443_2018}
{Texas Instruments}, ``{IWR1443} {Single}-{Chip} 76- to 81-{GHz} {mmWave} {Sensor} – {Data} {Interface} {With} {External} {Processor} {Over},'' 2018.

\bibitem{ronneberger_u-net_2015}
O.~Ronneberger, P.~Fischer, and T.~Brox, ``\BIBforeignlanguage{en}{U-{Net}: {Convolutional} {Networks} for {Biomedical} {Image} {Segmentation}},'' in \emph{\BIBforeignlanguage{en}{Medical {Image} {Computing} and {Computer}-{Assisted} {Intervention} – {MICCAI} 2015}}, N.~Navab, J.~Hornegger, W.~M. Wells, and A.~F. Frangi, Eds.\hskip 1em plus 0.5em minus 0.4em\relax Cham: Springer International Publishing, 2015, pp. 234--241.

\end{thebibliography}

\end{document}